\documentclass[11pt]{article}

\PassOptionsToPackage{table,dvipsnames}{xcolor}
\PassOptionsToPackage{hyperfootnotes=false}{hyperref}
\usepackage[preprint]{acl}

\usepackage{times}
\usepackage{latexsym}

\usepackage[T1]{fontenc}

\usepackage[utf8]{inputenc}

\usepackage{microtype}

\usepackage{inconsolata}

\usepackage{graphicx}

\usepackage{microtype}
\usepackage{hyperref}
\usepackage{url}
\usepackage{booktabs}

\usepackage[table,dvipsnames]{xcolor}

\usepackage[utf8]{inputenc}
\usepackage[T1]{fontenc}
\usepackage{microtype}

\usepackage{amsmath}
\usepackage{amssymb}
\usepackage{mathtools}
\usepackage{amsthm}
\usepackage{array}
\usepackage{tabularx}
\usepackage{ragged2e}
\usepackage{makecell}
\usepackage{booktabs}

\definecolor{headerblue}{RGB}{200,220,240}
\definecolor{textred}{RGB}{220,50,47}
\definecolor{ubgray}{RGB}{130,130,130}

\usepackage{multirow}
\usepackage{array}

\usepackage{caption}
\usepackage{fontawesome5}
\usepackage{graphicx}

\definecolor{sectiongray}{RGB}{240,240,240}
\definecolor{oursblue}{RGB}{230,238,248}
\definecolor{ubgray}{RGB}{130,130,130}

\newcolumntype{Y}{>{\RaggedRight\arraybackslash}X}

\newcolumntype{C}[1]{>{\centering\arraybackslash}m{#1}}

\usepackage{graphicx}
\usepackage{fontawesome5}
\usepackage{pifont}
\usepackage{tikz}

\usepackage{booktabs}
\usepackage{multirow}
\usepackage{makecell}
\usepackage{tabularx}
\usepackage{array}
\usepackage{ragged2e}
\usepackage{colortbl}

\usepackage[most]{tcolorbox}
\tcbuselibrary{skins, breakable, raster}
\usepackage{listings}

\usepackage{enumitem}
\usepackage{adjustbox}

\definecolor{darkblue}{rgb}{0, 0, 0.5}
\hypersetup{
  colorlinks=true,
  citecolor=darkblue,
  linkcolor=darkblue,
  urlcolor=darkblue
}

\title{Beyond Self-Reflection: An Execute--Distill--Verify Framework for Agentic Experience Learning}

\author{
\textbf{
Shiding Zhu$^{1}$\thanks{Equal contribution.} \quad
Yudi Qi$^{2}$\footnotemark[1] \quad
Yajie Wang$^{3}$\footnotemark[1] \quad
Jiaze Li$^{1}$ \quad
Chao Song$^{4}$
}
\\
\textbf{
Yaorui Shi$^{5}$ \quad
Yibo Miao$^{3}$ \quad
Hanqi Gao$^{6}$ \quad
Kai Zhang$^{2}$\thanks{\raggedright Corresponding author: \texttt{zhangkai@sz.tsinghua.edu.cn}. Additional emails: Shiding Zhu (\texttt{zhusd@zju.edu.cn}); Yudi Qi (\texttt{qyd24@mails.tsinghua.edu.cn}).}
}
\\[4mm]
{\normalfont\small $^{1}$Zhejiang University \quad}
{\normalfont\small $^{2}$Shenzhen International Graduate School, Tsinghua University} \\
{\normalfont\small $^{3}$Independent Researcher \quad}
{\normalfont\small $^{4}$Northwestern Polytechnical University} \\
{\normalfont\small $^{5}$University of Science and Technology of China \quad}
{\normalfont\small $^{6}$Shanghai Jiaotong University}
}

\hypersetup{
  pdftitle={Beyond Self-Reflection: An Execute-Distill-Verify Framework for Agentic Experience Learning},
  pdfauthor={Shiding Zhu, Yudi Qi, Yajie Wang, Jiaze Li, Chao Song, Yaorui Shi, Yibo Miao, Hanqi Gao, Kai Zhang},
  pdfsubject={Agentic experience learning},
  pdfkeywords={large language models, agents, experience learning, memory, EDV}
}

\begin{document}
\maketitle
\begin{abstract}
Experience learning lets large language model agents reuse lessons through external memory without updating model parameters. However, reflection can endorse execution errors and store them as future guidance, a failure mode we call the \textbf{Self-Confirmation Trap}. We propose \textbf{EDV}, an \textbf{Execute--Distill--Verify} framework for offline experience construction. Heterogeneous agents independently execute each task; a third-party distiller compares their trajectories and environmental feedback to extract candidate lessons; and the executors separately review each candidate, retaining only unanimously approved lessons. Evaluation uses a single solver retrieving from the frozen memory bank. Experiments on $\tau^2$-bench, Mind2Web, and MMTB show improved revisited-task performance, with additional evidence of within-domain transfer to held-out tasks on $\tau^2$-bench. On an annotated Retail subset, EDV reduces the Self-Confirmation Trap Rate from 11.0\% with RB-MaTTS to 5.0\%, indicating less retention of erroneous guidance.
\end{abstract}

\section{Introduction}

Large language model (LLM) agents learn from past interactions by
storing reusable lessons, skills, and strategies in external memory,
without updating model parameters
\citep{shinn2023reflexion,wang2023voyager,zhao2024expel,ouyang2025reasoningbank,zhang2025ace,zhang2026memrl}.
The value of this memory depends on the reliability and applicability
of its contents: preserving an incorrect action or interpretation as
guidance can carry the error beyond its original interaction
\citep{wang2026antipersistence}.

Self-reflection and feedback-guided refinement have been studied
across reasoning, code generation, and agent tasks
\citep{shinn2023reflexion,madaan2023selfrefine,chen2024selfdebug,gou2024critic,zhang-etal-2024-self-contrast,kamoi2024selfcorrection}.
However, self-correction can be unreliable in reasoning tasks without
external feedback
\citep{huang2023large,kamoi2024selfcorrection}.
Failed trajectories can yield useful lessons when their errors are
correctly diagnosed. The risk arises when reflection instead
endorses or rationalizes an erroneous action or decision and stores
it as reusable guidance. We call this failure mode the
\textbf{Self-Confirmation Trap}; subsequent retrieval may carry
the error into future decisions.

Our starting point is how people learn from one another and improve
through discussion and debate. Social cognitive theory highlights
learning from others' behavior \citep{bandura2001social}, while
argumentative reasoning emphasizes evaluating others' arguments to
uncover weaknesses missed individually \citep{mercier2011why}.
This motivates EDV to compare different agents' experiences, examine
disagreements, and review proposed lessons before retaining them as
shared guidance.

The central question is which candidate lessons are sufficiently
supported by task evidence to retain. Existing work uses reflection,
trajectory comparison, and separate curation roles to construct
experience
\citep{shinn2023reflexion,ouyang2025reasoningbank,zhang2025ace}.
Diverse solution sampling and multi-agent collaboration further
explore voting, comparison, and aggregation across alternative
responses
\citep{wang2022self,zhang-etal-2024-self-contrast,du2024improving,wang2024mixture,li2024more,ye2025x}.
Validation before acceptance also informs methods that update agent
behavior
\citep{jiang2026harnessevolve,liu2026sepm}.

Building on these ideas, we propose \textbf{EDV}, an
\textbf{Execute--Distill--Verify} framework for offline experience
construction. In \textbf{Execute}, heterogeneous agents independently
generate complete trajectories that can provide alternative approaches
and expose disagreements. In \textbf{Distill}, a third-party agent
compares these trajectories and environmental feedback to extract
candidate lessons with explicit applicability conditions, separating
lesson synthesis from the behavior being assessed. In \textbf{Verify},
the executors separately review each candidate against the collected
evidence; only unanimously approved lessons enter memory. The bank
is then frozen and retrieved by a single solver during evaluation.

We evaluate EDV on $\tau^2$-bench~\citep{barres2025tau},
Mind2Web~\citep{deng2023mind2web}, and MMTB~\citep{yu2025multi}.
Because agents may ignore or misinterpret condensed
experience~\citep{zhao2026faithful}, we examine both stored-experience
quality and subsequent task performance. Main experiments revisit
construction tasks through separate evaluations; additional
$\tau^2$-bench experiments assess within-domain transfer to held-out
tasks. On an annotated Retail subset, EDV lowers the
Self-Confirmation Trap Rate from 11.0\% with RB-MaTTS to 5.0\%.
Fixed-pool verification improves retained precision over random
retention at the same accepted count. Paired interventions further
show that removing or correcting retrieved erroneous lessons
reduces target-error recurrence on selected held-out cases.

Our contributions are threefold:
\begin{itemize}
    \item We characterize the \textbf{Self-Confirmation Trap}: execution errors endorsed during reflection and stored as persistent guidance.
    \item We propose \textbf{EDV}, combining heterogeneous execution, third-party distillation, and unanimous review before storage.
    \item We evaluate task performance and within-domain transfer, and connect experience-quality analysis with fixed-pool verification controls and paired memory interventions.
\end{itemize}

\section{Related Work}

\subsection{Memory-Based Experience Learning}

Memory-based agents adapt through external experience without updating the underlying language model.
\textit{Reflexion} stores verbal reflections derived from task feedback, and \textit{ExpeL} extracts reusable insights across training tasks
\citep{shinn2023reflexion,zhao2024expel}.
\textit{ReasoningBank} learns from self-judged successes and failures; its parallel \textit{MaTTS} variant compares multiple trajectories to synthesize memory
\citep{ouyang2025reasoningbank}.
\textit{ACE} separates generation, reflection, and curation, applying incremental updates to an evolving playbook
\citep{zhang2025ace}.
\textit{MemRL} uses environmental feedback to learn memory utility for retrieval
\citep{zhang2026memrl}.
\textit{MemEvolve} jointly evolves experience and memory architecture, while \textit{MemScope} uses execution traces, runtime memory state, and architecture code to guide localized repairs
\citep{zhang2025memevolve,su2026memscope}.
Related memory designs include \textit{Voyager}'s executable skill library,
\textit{A-MEM}'s linked memory notes, and \textit{HiAgent}'s
subgoal-based working-memory summaries
\citep{wang2023voyager,xu2025mem,hu2025hiagent}.
EDV builds on trajectory comparison and role separation, focusing on constructing and screening individual lessons within an offline memory-building pipeline.

\subsection{Reliability and Persistence of Experience}

Output refinement provides a related perspective on experience quality.
\textit{Self-Refine}, \textit{Self-Debugging}, \textit{CRITIC}, and
\textit{Self-Contrast} study revision using self-feedback, execution
results, external tools, or contrasting solution perspectives
\citep{madaan2023selfrefine,chen2024selfdebug,gou2024critic,zhang-etal-2024-self-contrast}.
Evaluations of intrinsic self-correction highlight the importance
of feedback quality and experimental conditions
\citep{huang2023large,kamoi2024selfcorrection}.
These studies examine output revision; EDV addresses which lessons
from execution should persist as future guidance.

Experience quality and its influence on behavior are distinct concerns.
\citet{zhao2026faithful} use controlled interventions to show that agents may ignore or misinterpret condensed experience.
\textit{More Memory, Worse Agents} analyzes how errors may be reproduced through abstraction, execution, and re-abstraction; its \textit{Anti-Persistence} design retains trajectories and outcomes while discarding task-specific guidance after use
\citep{wang2026antipersistence}.
\textit{JitMem} defers curation to read time, generates task-conditioned guidance from stored raw trajectories, and trains the curator using immediate task rewards
\citep{zhou2026jitmem}.
EDV retains distilled lessons as persistent memory and addresses their reliability before storage through cross-trajectory distillation and review.
Its Self-Confirmation Trap concerns execution errors endorsed during the transition from execution to memory.

\begin{figure*}[t]
    \centering
    \includegraphics[width=\linewidth]{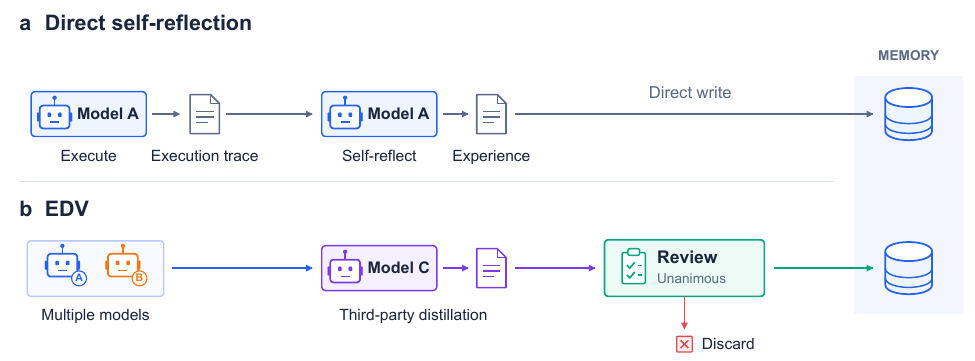}
    \vspace{-2mm}
    \caption{
    \textbf{Experience construction and memory admission.}
    Direct self-reflection uses the executor to extract lessons
    and writes them directly to memory.
    EDV separates heterogeneous execution, third-party distillation,
    and unanimous review before storage. The review stage is shown
    schematically; Figure~\ref{fig:picture_2} details the default
    trajectories and reviewer roles.
    }
    \label{fig:picture_1}
\end{figure*}

\begin{figure*}[t]
    \centering
    \includegraphics[width=\linewidth]{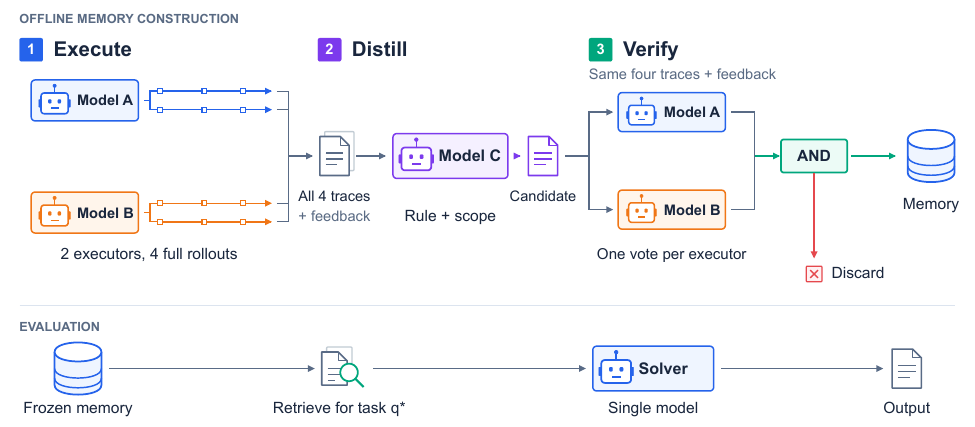}
    \vspace{-2mm}
    \caption{
    \textbf{EDV in the default configuration.}
    Two heterogeneous executors each produce two complete trajectories.
    Distiller C compares all four, and executors A and B separately
    review each proposed lesson using the same evidence.
    Each candidate receives two votes; both must approve it for storage.
    The bank is frozen after construction and used by a single
    evaluation-time solver. Task context and available policy/tool
    constraints are also provided to the distiller and reviewers.
    }
    \label{fig:picture_2}
\end{figure*}

\section{The Self-Confirmation Trap in Experience Learning}
\label{sec:self_confirmation}

A \textbf{Self-Confirmation Trap} occurs when an execution error
is endorsed or rationalized in a lesson stored as reusable experience.
As an illustration, a task prohibits using travel certificates to pay for
reservation changes. After this payment method is rejected, an agent
may attribute the failure to its action order and store advice to
reorder the steps and retry. The lesson still recommends the
prohibited payment method.

Identification requires two linked observations: task requirements,
rules, or environmental feedback establish an incorrect action or
decision in execution, and the stored lesson recommends or justifies
that behavior. Accurate failure records and corrective advice do
not meet this definition. This distinction separates recording an
unsuccessful attempt from preserving its error as future guidance.
We reserve the term Self-Confirmation Trap for errors already
present in execution that are endorsed during reflection and retained
as experience. New errors introduced during distillation also reduce
experience quality, but fall outside this failure mode.

\section{Method}

\subsection{Overview of EDV}

EDV separates independent execution, cross-trajectory distillation,
and review before storage. Figure~\ref{fig:picture_1} contrasts it
with direct self-reflection; Figure~\ref{fig:picture_2} shows the
default workflow. A single solver retrieves the resulting frozen
memory bank during evaluation.

\subsection{Execute: Independent Execution with Different Models}

Execution agents independently complete the same task from its
initial setup, without exchanging intermediate results or selecting
between proposed actions at individual steps. Distillation begins
only after all runs finish.

For task $q$, let
$\mathcal{A}_{\mathrm{exec}}=\{A_1,\ldots,A_N\}$ denote agents
using different foundation models. Agent $A_i$ produces $R_i\geq1$
complete trajectories $\tau_{i,r}$. Their collection and total size are
\begin{equation}
\begin{aligned}
\mathcal{T}(q)
&=\bigl\{\tau_{i,r}\bigr\}_{1\leq i\leq N,\,1\leq r\leq R_i},\\
L&=\sum_{i=1}^{N}R_i.
\end{aligned}
\end{equation}
By default, two heterogeneous executors each produce two runs
($N=2$, $R_1=R_2=2$, $L=4$), with a third model reserved for
distillation.

\subsection{Distill: Extracting Lessons from Multiple Trajectories}

In default EDV, the distiller uses a model that did not execute
this task. Its inputs include all trajectories, the task description,
available rules, tool definitions, environmental feedback,
and retrieved experience.
By comparing actions and outcomes across trajectories, it proposes
lessons about which strategies to use, which mistakes to avoid,
and when the advice applies.

We denote this shared information by $\mathcal{B}(q)$.
The distiller produces a set of candidate lessons:
\begin{equation}
\mathcal{E}_{\mathrm{cand}}(q)
=D\bigl(\mathcal{B}(q)\bigr)
=\{e_1,\ldots,e_J\},
\end{equation}
where $J$ is the number of candidates.

The distiller first selects the strongest available trajectory,
prioritizing compliance, then accuracy, then efficiency. It then
compares this trajectory with successes and failures in the others
to extract evidence-supported candidate lessons, checking them
against task rules and tool requirements.
The proposed lessons are reviewed in the next stage.

\subsection{Verify: Reviewing Lessons Before Storage}

The execution agents independently review the candidate set,
assigning a decision to each item. Every reviewer receives
the same evidence $\mathcal{B}(q)$,
including all collected trajectories.
It checks whether the lesson is supported by the evidence,
follows the applicable rules, and gives specific, useful advice
within its stated scope. Incorrect, unsupported, overly general,
or redundant lessons can be rejected.

Each reviewer casts one vote per candidate. A lesson is retained
only if every reviewer approves it.
Thus, in the default setting, each candidate receives two votes
based on four trajectories. A single rejection excludes it.

Let $V_i(e;\mathcal{B}(q))$ be the vote from reviewer $A_i$:
$1$ means approval and $0$ means rejection.
The retained lessons are
\begin{equation}
\begin{aligned}
\mathcal{E}_{\mathrm{keep}}(q)
&=\bigl\{e\in\mathcal{E}_{\mathrm{cand}}(q):\\
&\quad V_i(e;\mathcal{B}(q))=1,\ \forall i\bigr\}.
\end{aligned}
\end{equation}

This review uses existing evidence and does not rerun the task
to test each candidate.

\subsection{Storing and Reusing Experience}

During offline construction, execution agents can retrieve lessons
accumulated from earlier tasks. Approved new lessons are added to
the memory bank.
Once offline construction ends, the bank remains unchanged
during evaluation. For each evaluation task, a single solver
retrieves relevant lessons and uses them as additional context.
Its new execution records are not added to the bank.
Implementation details are provided in
Appendix~\ref{appendix:implementation_details}.

\begin{table}[t]
\centering
\captionsetup{justification=raggedright,singlelinecheck=false}
\caption{
Pass@1 results on $\tau^2$-bench across the Airline, Retail, and Telecom domains. Higher values indicate better task-completion performance, and Avg.\ reports the average score across the three domains.
}
\label{tab:tau2_main}

\renewcommand{\arraystretch}{1.15}
\setlength{\tabcolsep}{3pt}
\small

\begin{tabular*}{\columnwidth}{
@{\extracolsep{\fill}}lrrrr@{}
}
\toprule
\textbf{Method}
& \textbf{Airline}
& \textbf{Retail}
& \textbf{Telecom}
& \textbf{Avg.} \\
\midrule
NM       & 61.5 & 82.2 & 85.5 & 76.4 \\
DSR      & 63.5 & 82.9 & 86.6 & 77.7 \\
RB       & 66.0 & 83.3 & 87.7 & 79.0 \\
RB-MaTTS & 68.5 & 86.0 & 92.4 & 82.3 \\
\midrule
\textbf{EDV}
& \textbf{72.0}
& \textbf{88.6}
& \textbf{99.1}
& \textbf{86.6} \\
\bottomrule
\end{tabular*}

\end{table}

\begin{table*}[t]
\centering
\captionsetup{justification=raggedright,singlelinecheck=false}
\caption{
Revisited-task results on the Cross-Task, Cross-Website, and Cross-Domain subsets of Mind2Web. Higher values indicate better performance for all metrics. Metric definitions are provided in Appendix~\ref{appendix:benchmark_details}.
}
\label{tab:mind2web_main}

\renewcommand{\arraystretch}{1.15}
\setlength{\tabcolsep}{3.5pt}
\footnotesize

\begin{tabular*}{\textwidth}{
@{\extracolsep{\fill}}
l
rrrr
rrrr
rrrr
@{}
}
\toprule
\multirow{2}{*}{\textbf{Method}}
& \multicolumn{4}{c}{\textbf{Cross-Task}}
& \multicolumn{4}{c}{\textbf{Cross-Website}}
& \multicolumn{4}{c}{\textbf{Cross-Domain}} \\
\cmidrule(lr){2-5}
\cmidrule(lr){6-9}
\cmidrule(lr){10-13}

& \textbf{EA}
& \textbf{AF1}
& \textbf{SSR}
& \textbf{SR}
& \textbf{EA}
& \textbf{AF1}
& \textbf{SSR}
& \textbf{SR}
& \textbf{EA}
& \textbf{AF1}
& \textbf{SSR}
& \textbf{SR} \\
\midrule

NM
& 39.92 & 59.14 & 34.19 & 3.57
& 35.03 & 49.59 & 29.13 & 2.82
& 36.24 & 52.00 & 31.52 & 2.74 \\

DSR
& 40.21 & 59.62 & 34.88 & 3.69
& 35.74 & 50.03 & 29.86 & 3.05
& 36.98 & 52.66 & 32.31 & 2.96 \\

RB
& 41.36 & 60.41 & 36.02 & 3.95
& 36.42 & 50.79 & 30.72 & 3.31
& 37.95 & 53.58 & 33.53 & 3.24 \\

RB-MaTTS
& 44.87 & 61.78 & 39.61 & 4.31
& 41.26 & 53.88 & 34.54 & 4.12
& 40.92 & 54.96 & 36.84 & 4.08 \\

\midrule

\textbf{EDV}
& \textbf{48.62}
& \textbf{63.10}
& \textbf{43.17}
& \textbf{4.76}
& \textbf{44.79}
& \textbf{55.64}
& \textbf{36.56}
& \textbf{4.76}
& \textbf{43.39}
& \textbf{56.38}
& \textbf{39.57}
& \textbf{4.71} \\

\bottomrule
\end{tabular*}

\end{table*}

\begin{table*}[t]
\centering
\captionsetup{justification=raggedright,singlelinecheck=false}
\caption{
Revisited-task results on MMTB. Overall is reported from its separate evaluation; the remaining columns report action-level, multi-action, and trajectory-level metrics. Higher values indicate better performance for all metrics. Metric definitions are provided in Appendix~\ref{appendix:benchmark_details}.
}
\label{tab:mmtb_main}

\renewcommand{\arraystretch}{1.15}
\setlength{\tabcolsep}{3pt}
\footnotesize

\begin{tabular*}{\textwidth}{
@{\extracolsep{\fill}}
l
r
rrr
rrr
rr
@{}
}
\toprule
\multirow{2}{*}{\textbf{Method}}
& \textbf{Overall}
& \multicolumn{3}{c}{\textbf{Action-Level}}
& \multicolumn{3}{c}{\textbf{Multi-Action}}
& \multicolumn{2}{c}{\textbf{Trajectory}} \\
\cmidrule(lr){2-2}
\cmidrule(lr){3-5}
\cmidrule(lr){6-8}
\cmidrule(lr){9-10}

& \textbf{All}
& \(A_{\mathrm{sgl}}\)
& \(A_{\mathrm{cht}}\)
& \(A_{\mathrm{clr}}\)
& \(A^{\mathrm{P}}\)
& \(A^{\mathrm{S}}\)
& \(A^{\mathrm{S+P}}\)
& \textbf{Opt. Path}
& \textbf{Acc. Prog.} \\
\midrule

NM
& 51.56
& 56.64 & 83.20 & 36.72
& 42.04 & 12.50 & 9.52
& 29.05 & 29.31 \\

DSR
& 52.48
& 57.03 & 83.59 & 38.28
& 43.31 & 18.75 & 14.29
& 30.41 & 34.76 \\

RB
& 53.67
& 56.47 & 83.98 & 39.84
& 42.04 & 25.00 & 20.73
& 32.22 & 40.03 \\

RB-MaTTS
& 55.72
& 58.59 & 84.10 & 42.58
& 47.13 & 31.25 & 25.00
& 35.67 & 46.81 \\

\midrule

\textbf{EDV}
& \textbf{58.10}
& \textbf{60.94}
& \textbf{84.38}
& \textbf{46.09}
& \textbf{51.59}
& \textbf{37.50}
& \textbf{29.76}
& \textbf{39.83}
& \textbf{50.42} \\

\bottomrule
\end{tabular*}

\end{table*}

\section{Experiments}
\subsection{Experimental Setup}

\paragraph{Benchmarks and Metrics.}
We evaluate EDV on $\tau^2$-bench for policy-constrained service
interactions~\citep{barres2025tau}, Mind2Web for web
navigation~\citep{deng2023mind2web}, and MMTB for tool-use
tasks~\citep{yu2025multi}.
For $\tau^2$-bench, we report Pass@1 in the Airline, Retail,
and Telecom domains and their unweighted average.
For Mind2Web, we report element accuracy, action F1,
step success rate, and task success rate.
For MMTB, we report the overall score and component-level metrics.
Benchmark configurations and metric definitions are provided in
Appendix~\ref{appendix:benchmark_details}.

\paragraph{Evaluation Protocols.}
Main experiments revisit construction tasks through separate
evaluation executions using frozen memory banks. During construction,
methods can retrieve experience from earlier tasks; evaluation
trajectories are never stored. Banks are separate for each
$\tau^2$-bench domain, each Mind2Web subset, and MMTB.
Mind2Web's Cross-Task, Cross-Website, and Cross-Domain names identify
the original subsets, not held-out splits for memory construction.
Section~\ref{sec:unseen_tasks} separately evaluates within-domain
transfer using disjoint construction and evaluation task sets.

\paragraph{Baselines.}
We compare EDV with four baselines.
\textbf{No Memory (NM)} solves tasks without stored experience.
\textbf{Direct Self-Reflection (DSR)} generates one trajectory per
construction task, then uses the executor to extract and directly
store lessons from that trajectory.
\textbf{ReasoningBank (RB)} also generates one trajectory per task
and extracts reusable reasoning lessons using model-assessed
success or failure~\citep{ouyang2025reasoningbank}.
\textbf{RB-MaTTS} uses the parallel variant, generating four
trajectories per task with the same model and comparing them to
construct experience.
DSR, RB, and RB-MaTTS use MiniMax-M2.1 for memory construction.
RB and RB-MaTTS are adapted to the construction and frozen-memory
evaluation protocol described above.

\paragraph{Implementation Details.}
The main experiments use MiniMax-M2.1 as the common solver and
share retrieval settings across memory-based methods. EDV uses
MiniMax-M2.1, Mimo-V2-Flash, and GLM-4.7-FP8 in the construction
workflow of Section~4. Construction uses task rules, execution
trajectories, and tool feedback, without reference answers or
ground-truth evaluation labels. Prompt excerpts and implementation
settings appear in Appendices~\ref{appendix:prompts}
and~\ref{appendix:implementation_details}.
Supplementary analyses in Appendix~\ref{sec:additional} use
GLM-4.7-FP8 as the solver.

For construction, Execute and Distill use temperature 1.0 for
MiniMax-M2.1 and GLM-4.7-FP8, and 0.3 for Mimo-V2-Flash.
Verify uses 0.0 for all models. Generations are capped at 8,192
output tokens; temperature zero alone does not guarantee
deterministic backend responses.

\subsection{Main Results}
\label{sec:main_results}

Tables~\ref{tab:tau2_main}, \ref{tab:mind2web_main}, and
\ref{tab:mmtb_main} report revisited-task results under the
protocol in Section~5.1. EDV outperforms all compared methods
across the reported metrics.

On $\tau^2$-bench, EDV achieves an average Pass@1 of 86.6\%,
exceeding No Memory and RB-MaTTS by 10.2 and 4.3 percentage
points, respectively. The largest gain over RB-MaTTS occurs
on Telecom, where Pass@1 increases from 92.4\% to 99.1\%.

On Mind2Web, EDV improves all four metrics across the three
subsets compared with RB-MaTTS, with gains of 2.02--3.56
percentage points in step success rate and 0.45--0.64 points
in task success rate.

On MMTB, EDV raises the overall score from 55.72 to 58.10
compared with RB-MaTTS, while improving parallel, sequential,
and mixed tool-use scores by 4.46, 6.25, and 4.76 percentage
points, respectively.

\subsection{Self-Confirmation Trap and Experience Quality Analysis}
\label{sec:trap_quality}

We descriptively compare stored errors, verification selectivity,
and subsequent behavior. Appendix~\ref{appendix:experience_analysis}
gives the protocols and reporting limits.

\paragraph{Human Evaluation.}
We compare stored experience from DSR, RB, RB-MaTTS, and EDV on a
common set of \textsc{Retail} tasks with at least one failed execution
(Table~\ref{tab:trap_quality}). Three annotators assess each item
against its source evidence with method identities hidden.
Trap labels follow Section~\ref{sec:self_confirmation} and are
resolved by majority vote.

\paragraph{Evaluation Metrics.}
The Self-Confirmation Trap Rate (SCTR) is
\begin{equation}
\mathrm{SCTR}
=\frac{N_{\mathrm{trap}}}{N_{\mathrm{annotated}}}\times100\%,
\label{eq:sctr}
\end{equation}
where the numerator counts trap-labeled items and the denominator
all annotated items. Annotators also rate correctness,
actionability, and specificity on 1--5 scales, assessing consistency
with evidence, concrete guidance, and clear applicability conditions.
Scores are averaged across annotators and then items; protocols
are in Appendix~\ref{appendix:experience_analysis}.

Table~\ref{tab:trap_quality} shows an SCTR of 5.0\% for EDV,
compared with 22.0\% for DSR, 16.0\% for RB, and 11.0\% for
RB-MaTTS. Relative to RB-MaTTS, this is a 6.0-percentage-point
reduction (54.5\% relative). Correctness rises from 4.0 to 4.5
and specificity from 4.0 to 4.3, while actionability remains 4.2.

\begin{table}[t]
\centering
\captionsetup{justification=raggedright,singlelinecheck=false}
\caption{
Stored-experience quality on a common set of \textsc{Retail} tasks.
SCTR is a percentage. Corr., Act., and Spec. denote correctness,
actionability, and specificity on a 1--5 scale (higher is better).
}
\label{tab:trap_quality}
\renewcommand{\arraystretch}{1.10}
\setlength{\tabcolsep}{1.5pt}
\footnotesize
\begin{tabular*}{\columnwidth}{@{\extracolsep{\fill}}lcccc@{}}
\toprule
Method & SCTR $\downarrow$ & Corr. & Act. & Spec. \\
\midrule
DSR & 22.0 & 3.5 & 3.9 & 3.4 \\
RB & 16.0 & 3.8 & 4.0 & 3.7 \\
RB-MaTTS & 11.0 & 4.0 & \textbf{4.2} & 4.0 \\
\midrule
\textbf{EDV} & \textbf{5.0} & \textbf{4.5} & \textbf{4.2} & \textbf{4.3} \\
\bottomrule
\end{tabular*}
\end{table}

\paragraph{Verification Selectivity.}
We compare selection rules on the same pre-verification candidate
pool, using human correctness labels for accepted and rejected
items. The rules include keeping all candidates, either reviewer
alone, unanimous approval, and random retention matched to EDV's
accepted count within each task.
Table~\ref{tab:verification_selectivity} reports precision among
retained items, recall of correct candidates, rejection recall for
incorrect candidates, and the fraction of all candidates accepted.

EDV retains 68.0\% of candidates with 96.5\% precision, compared
with 80.0\% for matched random retention. At the same accepted
count, it retains more correct candidates (82.0\% versus 68.0\%
recall) and rejects more errors (88.0\% versus 32.0\%). This
comparison supports selective filtering beyond reducing memory
volume alone. Relative to either reviewer alone, unanimity raises
precision from 90.0--91.7\% to 96.5\% and error rejection from
60.0--68.0\% to 88.0\%, but lowers correct-memory recall from
88.0--90.0\% to 82.0\%. Verification therefore trades the loss
of some correct advice for a cleaner retained pool.

\begin{table}[t]
\centering
\captionsetup{justification=raggedright,singlelinecheck=false}
\caption{
Verification selectivity on a fixed candidate pool.
Prec., Rec., Rej., and Accept. denote retained precision,
correct-memory recall, error-rejection recall, and acceptance rate,
all in percent. Random retention matches EDV's accepted count
within each task.
}
\label{tab:verification_selectivity}
\renewcommand{\arraystretch}{1.10}
\setlength{\tabcolsep}{2pt}
\small
\begin{tabular*}{\columnwidth}{@{\extracolsep{\fill}}lcccc@{}}
\toprule
Selection rule & Prec. & Rec. & Rej. & Accept. \\
\midrule
Keep all & 80.0 & 100.0 & 0.0 & 100.0 \\
A only & 90.0 & 90.0 & 60.0 & 80.0 \\
B only & 91.7 & 88.0 & 68.0 & 76.8 \\
Random (matched) & 80.0 & 68.0 & 32.0 & 68.0 \\
\midrule
\textbf{A AND B (EDV)} & \textbf{96.5} & 82.0 & \textbf{88.0} & 68.0 \\
\bottomrule
\end{tabular*}
\end{table}

\paragraph{Behavioral Effects of Erroneous Experience.}
On held-out \textsc{Retail} tasks, we select cases where ordinary
retrieval returns a human-confirmed trap. We retain the lesson,
remove it without backfilling, or correct its erroneous claim
without adding a target-task answer. Other retrieved items and
their order remain fixed. All conditions use MiniMax-M2.1 in fresh
sessions with matched execution settings; blinded annotators judge
recurrence of the prespecified target error.

Across 60 executions per condition
(Table~\ref{tab:memory_intervention}), removing the lesson reduces
target-error recurrence from 50.0\% to 30.0\% and raises task
success from 40.0\% to 55.0\%. Correction further lowers recurrence
to 20.0\% and raises success to 65.0\%: improvements of 30.0 and
25.0 percentage points, respectively, over the original lesson.
These interventions indicate that erroneous guidance can influence
subsequent behavior, with correction outperforming removal on
these selected cases. Together with the quality and selectivity
analyses, this finding supports checking the content of reusable
experience before it can influence subsequent decisions.

\begin{table}[t]
\centering
\captionsetup{justification=raggedright,singlelinecheck=false}
\caption{
Behavioral effects of erroneous experience through paired
interventions on a retrieved erroneous lesson.
$n$ is the number of task executions per condition.
Error is target-error recurrence (\%); Success is task success (\%).
The same task cases are evaluated under all three conditions.
}
\label{tab:memory_intervention}
\renewcommand{\arraystretch}{1.10}
\setlength{\tabcolsep}{3pt}
\small
\begin{tabular*}{\columnwidth}{@{\extracolsep{\fill}}lccc@{}}
\toprule
Memory condition & $n$ & Error $\downarrow$ & Success $\uparrow$ \\
\midrule
Original lesson & 60 & 50.0 & 40.0 \\
Lesson removed & 60 & 30.0 & 55.0 \\
Lesson corrected & 60 & 20.0 & 65.0 \\
\bottomrule
\end{tabular*}
\end{table}

\begin{table*}[t]
\centering
\captionsetup{justification=raggedright,singlelinecheck=false}
\caption{
Ablation and token-usage comparisons on revisited \textsc{Retail}
tasks in $\tau^2$-bench. All configurations use MiniMax-M2.1
for evaluation. Construction and evaluation token usage are
normalized to DSR and No Memory, respectively.
Homogeneous EDV uses the indicated backbone for all construction
roles. Dual Review requires unanimous approval of candidate
experience by the two models other than the distillation model.
}
\label{tab:ablation}
\renewcommand{\arraystretch}{1.10}
\setlength{\tabcolsep}{4pt}
\small
\begin{tabular*}{\textwidth}{@{\extracolsep{\fill}}lrrr@{}}
\toprule
Method
& Construction Tokens $\downarrow$
& Evaluation Tokens $\downarrow$
& Pass@1 (\%) $\uparrow$ \\
\midrule
\multicolumn{4}{@{}l}{\textit{A. Reference methods}} \\
No Memory & 0.00 & 1.00 & 82.2 \\
DSR & 1.00 & 0.98 & 82.9 \\
RB & 1.10 & 0.95 & 83.3 \\
RB-MaTTS & 1.63 & 0.89 & 86.0 \\
\midrule
\multicolumn{4}{@{}l}{\textit{B. Multi-trajectory construction}} \\
Homogeneous Multi-Traj. & 1.56 & 0.92 & 84.9 \\
Heterogeneous Multi-Traj. & 1.56 & 0.93 & 85.9 \\
\midrule
\multicolumn{4}{@{}l}{\textit{C. EDV variants}} \\
EDV w/o Verify & 1.75 & 0.85 & 87.1 \\
Homogeneous EDV (MiniMax-M2.1) & 1.85 & 0.84 & 86.6 \\
Homogeneous EDV (Mimo-V2-Flash) & 1.85 & 0.81 & 86.8 \\
Homogeneous EDV (GLM-4.7-FP8) & 1.85 & 0.82 & 87.1 \\
\midrule
\multicolumn{4}{@{}l}{\textit{D. Additional control}} \\
RB-MaTTS + Dual Review & 1.86 & 0.87 & 86.4 \\
\midrule
\textbf{EDV} & 1.85 & \textbf{0.79} & \textbf{88.6} \\
\bottomrule
\end{tabular*}
\end{table*}

\subsection{Ablation and Token Usage}
\label{sec:ablation}

Table~\ref{tab:ablation} compares construction configurations on
revisited \textsc{Retail} tasks, using MiniMax-M2.1 as the fixed
solver. All multi-trajectory variants generate four complete runs
per construction task. Additional construction details are in
Appendix~\ref{appendix:construction_controls}.

\paragraph{Multi-Trajectory Construction.}
Homogeneous multi-trajectory construction achieves 84.9\% Pass@1,
compared with 82.9\% for single-trajectory DSR. Heterogeneous
construction reaches 85.9\%, improving by 1.0 percentage point at
the same reported normalized construction token usage of 1.56.
Both variants use four trajectories per task. This comparison
does not fully control distillation and role allocation, so the
difference cannot be attributed to model diversity alone.

\paragraph{Effect of Verification.}
Adding verification to EDV raises Pass@1 from 87.1\% to 88.6\%.
Both variants use a distiller that did not participate in execution;
the 1.5-point gain supports reviewing candidates before memory
insertion within this workflow.

\paragraph{Model Heterogeneity in the Full Workflow.}
Homogeneous EDV retains separate execution, distillation, and review
roles using one backbone throughout. Its MiniMax-M2.1, Mimo-V2-Flash,
and GLM-4.7-FP8 variants achieve 86.6\%, 86.8\%, and 87.1\%,
respectively. Heterogeneous EDV reaches 88.6\%, 1.5 points above
the strongest homogeneous variant. This compares complete workflows;
it does not isolate model diversity from role assignment.

\paragraph{Token Usage and Performance.}
We count input and output tokens across all relevant model calls,
normalizing construction usage to DSR and evaluation usage to
No Memory. Relative to RB-MaTTS, EDV uses 13.5\% more construction
tokens (1.63 to 1.85) and 11.2\% fewer evaluation tokens
(0.89 to 0.79), while Pass@1 rises from 86.0\% to 88.6\%.

RB-MaTTS + Dual Review applies the same unanimous candidate-approval
rule, using the two models other than the distiller as reviewers.
It achieves 86.4\% Pass@1 at normalized construction usage of 1.86;
EDV achieves 88.6\% at 1.85, a 2.2-point advantage at similar
measured construction token usage.

\subsection{Within-Domain Generalization}
\label{sec:unseen_tasks}
Our main experiments evaluate experience reuse on revisited tasks.
We further examine whether the constructed experience benefits
held-out tasks within the Airline, Retail, and Telecom domains
of $\tau^2$-bench.

\paragraph{Evaluation Protocol.}
Within each domain, we randomly divide tasks into two approximately
equal-sized, disjoint sets, keeping all trajectories associated
with each task in the same set. The partitions are shared across
methods. We use two-fold evaluation: each memory-based method
constructs experience on one set and evaluates on the other, then
reverses their roles. Each fold starts with an empty memory bank,
which is frozen after construction. Memories are not shared across
folds or domains, and evaluation trajectories are never stored.
Every evaluation task is therefore excluded from the construction
of the memory used to solve it.

All methods use MiniMax-M2.1 as the evaluation-time solver, and
memory-based methods share the same retrieval configuration.
We repeat the complete construction and evaluation procedure
three times with fixed task partitions. For each run, domain-level
Pass@1 is computed by pooling held-out evaluations from both folds.
We report mean scores across the three runs and the unweighted
average across domains.

\paragraph{Results.}
Table~\ref{tab:unseen_tasks} shows that EDV achieves the highest
Pass@1 among the compared methods in all three domains.
Its average score reaches 85.1\%, compared with 76.4\% for
No Memory and 81.1\% for RB-MaTTS.
Relative to RB-MaTTS, EDV improves Pass@1 by 4.0, 2.0, and
5.8 percentage points on Airline, Retail, and Telecom, respectively.
These results show that the benefits of EDV-constructed experience
extend to tasks excluded from memory construction, supporting
its reuse across tasks within the same domain.

\begin{table}[t]
\centering
\captionsetup{
    justification=raggedright,
    singlelinecheck=false
}
\caption{
Within-domain generalization on held-out tasks in $\tau^2$-bench
under two-fold evaluation. Scores are mean Pass@1 (\%) over three
complete runs with fixed task partitions.
Avg.\ is the unweighted average across the three domains.
}
\label{tab:unseen_tasks}

\renewcommand{\arraystretch}{1.10}
\setlength{\tabcolsep}{2pt}
\small

\begin{tabular*}{\columnwidth}{
    @{\extracolsep{\fill}}lcccc@{}
}
\toprule
Method & Airline & Retail & Telecom & Avg. \\
\midrule
NM       & 61.5 & 82.2 & 85.5 & 76.4 \\
DSR      & 61.7 & 82.3 & 85.4 & 76.5 \\
RB       & 64.0 & 82.9 & 86.7 & 77.9 \\
RB-MaTTS & 67.0 & 85.0 & 91.4 & 81.1 \\
\midrule
\textbf{EDV}
& \textbf{71.0}
& \textbf{87.0}
& \textbf{97.2}
& \textbf{85.1} \\
\bottomrule
\end{tabular*}

\end{table}

\section{Conclusion}
\label{sec:conclusion}

EDV combines heterogeneous execution, third-party distillation,
and unanimous review to address execution errors endorsed as
reusable guidance. A single solver retrieves from the frozen
memory bank.

Experiments on $\tau^2$-bench, Mind2Web, and MMTB show improved
performance on revisited tasks, with additional evidence of
within-domain transfer to held-out $\tau^2$-bench tasks.
Retail analyses show a lower stored Self-Confirmation Trap Rate and
higher retained precision under verification, at the cost of
rejecting some correct candidates. Paired interventions show that
removing or correcting erroneous guidance reduces target-error
recurrence on selected held-out cases. These findings support
reviewing lessons before storage and assessing their behavioral
effects.

\section{Limitations}
\label{sec:limitations}

\paragraph{Reliability of Model-Based Review.}
EDV relies on models to interpret execution evidence. Different
models may share errors, and execution agents acting as reviewers
may retain their original misconceptions. Even the best available
trajectory can be incorrect, and tool acceptance alone does not
establish compliance with task rules. Verification reviews existing
evidence without testing candidate lessons through new executions.
Unanimous approval therefore does not guarantee correctness and
may also reject useful advice when one reviewer disagrees.

\paragraph{Continual Experience Use.}
EDV uses a fixed three-stage workflow and freezes memory during
evaluation. The experiments therefore do not assess continual
online learning or long-term memory maintenance. Dynamic role
assignment, conflicting or outdated lessons, memory growth, and
experience revision as task rules or tools change remain open
directions.

\paragraph{Scope of Experience-Quality Evidence.}
Our detailed quality and intervention analyses focus on selected
Retail cases. Extending these analyses to other domains and longer
interaction histories would help assess how broadly the observed
benefits of experience verification generalize. Missing item counts
and repeat-level results limit uncertainty estimates for these
descriptive comparisons (Appendix~\ref{appendix:experience_analysis}).

\bibliography{custom}
\clearpage
\appendix

\twocolumn[{
\begin{minipage}{\textwidth}
\centering
\includegraphics[width=\textwidth]{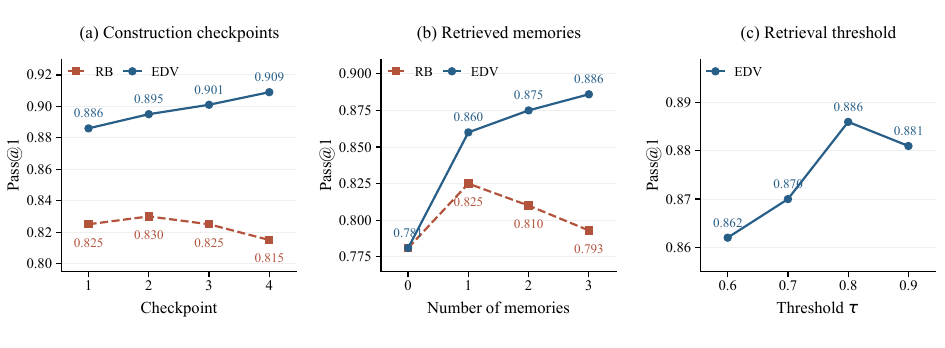}
\captionsetup{hypcap=false}
\captionof{figure}{Supplementary Pass@1 analyses with GLM-4.7-FP8 as solver:
(a) successive memory-construction checkpoints, (b) zero to three retrieved
memories, and (c) four retrieval thresholds. Checkpoints index external-memory
construction, not model-parameter updates. The panels report aggregate scores
under their respective settings; uncertainty across repeated runs is not shown.}
\label{fig:appendix_sweeps}
\vspace{10pt}
\end{minipage}
}]
\section{Additional Experimental Results}
\label{sec:additional}

These supplementary analyses use GLM-4.7-FP8 as the
evaluation-time solver, whereas the main experiments use
MiniMax-M2.1. We report the aggregate scores shown in the plots
and restrict interpretation to the displayed settings. These
curves do not establish statistical significance or performance
on new domains.

\subsection{Memory-Construction Checkpoints}
Figure~\ref{fig:appendix_sweeps}(a) compares EDV and ReasoningBank
(RB) at four successive construction checkpoints. EDV increases
from 88.6\% to 90.9\%, while RB increases from 82.5\% to 83.0\%
and then decreases to 81.5\%. The EDV--RB differences are 6.1,
6.5, 7.6, and 9.4 percentage points, respectively. The displayed
EDV scores increase across these checkpoints; four aggregate
measurements alone do not establish convergence or stability
across independent runs. Model parameters are not updated in
this memory-construction workflow.

\subsection{Number of Retrieved Memories}
Both methods score 78.1\% without retrieved memories
(Figure~\ref{fig:appendix_sweeps}(b)). EDV reaches 86.0\%, 87.5\%,
and 88.6\% with one, two, and three items, respectively. RB
reaches 82.5\% with one item, then decreases to 81.0\% and 79.3\%.
Thus, additional EDV memories are associated with higher scores
within the tested range. The curves do not isolate whether RB's
decline arises from incorrect advice, relevance, or other effects
of additional context, and do not establish a trend beyond three
retrieved items.

\subsection{Retrieval-Threshold Sensitivity}
Figure~\ref{fig:appendix_sweeps}(c) varies the memory-retrieval
threshold $\tau$. Scores are 86.2\%, 87.0\%, 88.6\%, and 88.1\%
at thresholds 0.6, 0.7, 0.8, and 0.9, respectively. A threshold
of 0.8 gives the highest score among these four tested settings.
This observation is specific to this sweep; it neither identifies
a generally optimal threshold nor establishes the default
retrieval setting of the main experiments.

\section{Implementation Details}
\label{appendix:implementation_details}

\subsection{Construction Roles and Generation Settings}

Default EDV uses two heterogeneous executors, each producing two
complete task trajectories ($N=2$, $L=4$). A third model performs
distillation after the executions finish. The two executors then
review the candidate memories independently. Roles remain fixed
within a task, and no agent selects between competing actions
at individual execution steps.

For construction, Execute and Distill use temperature 1.0 for
MiniMax-M2.1 and GLM-4.7-FP8, and 0.3 for Mimo-V2-Flash. Verify
uses temperature 0.0 for all models. Each generation is capped at
8,192 output tokens. Temperature zero does not by itself guarantee
deterministic responses from the serving backend. Additional
construction variants are described in
Appendix~\ref{appendix:construction_controls}.

\subsection{Memory Construction and Evaluation}

Memory banks are maintained separately for each independent
construction/evaluation partition. Execution agents can retrieve
experience accumulated from earlier construction tasks. Approved
new lessons are added to the bank. After construction, the bank
is frozen: evaluation trajectories are neither stored nor used to
revise experience.

Memory-based methods share the retrieval configuration within
each evaluation setting. The main experiments use MiniMax-M2.1
as the evaluation-time solver; the supplementary plots in
Appendix~\ref{sec:additional} use GLM-4.7-FP8. Retrieved lessons
provide additional context without updating model parameters.
Memory retrieval is separate from Mind2Web's DOM-element candidate
retrieval; the latter's Top-10 limit is not a memory top-$k$ setting.

\subsection{Task Separation and Reporting Scope}

Main results revisit construction tasks using separate evaluation
executions. Memory banks are separate for each $\tau^2$-bench domain,
each Mind2Web subset, and MMTB. The held-out experiment in
Section~\ref{sec:unseen_tasks} additionally separates construction
and evaluation tasks within each fold, with no sharing of memories
across folds or domains.

The reported protocol and aggregate results do not include a
complete record of retrieval parameters, task ordering, random
seeds, or model-role allocations across tasks. These omissions
limit exact reproduction of the memory banks and retrieved contexts;
no unreported settings are inferred from the sensitivity plots.

\section{Benchmark Specifications and Evaluation Protocols}
\label{appendix:benchmark_details}

\subsection{\texorpdfstring{$\tau^2$-Bench}{tau2-Bench}}

$\tau^2$-bench evaluates agents in policy-constrained service
interactions~\citep{barres2025tau}. We use the Airline, Retail,
and Telecom domains and follow the benchmark's task-success
evaluation. Pass@1 reports success for a single task execution,
rather than whether any of several attempts succeeds. Domain
scores are reported separately, and Avg.\ is their unweighted
mean. Construction rollouts are distinct from evaluation attempts.
The held-out two-fold protocol is described in
Section~\ref{sec:unseen_tasks}.

\subsection{Mind2Web}

Mind2Web provides annotated web tasks with step-level target
elements and actions~\citep{deng2023mind2web}. Cross-Task,
Cross-Website, and Cross-Domain are names of its original
evaluation subsets. In our main memory experiment, construction
and evaluation revisit tasks within each subset; these names do
not imply that evaluation tasks were held out from memory
construction.

\paragraph{Metrics.}
\textbf{Element Accuracy (EA)} indicates whether the predicted
element matches an annotated target. \textbf{Action F1 (AF1)} compares tokens in the
predicted action and value against the reference. For token sets
$T_p$ and $T_r$, precision and recall are
$|T_p\cap T_r|/|T_p|$ and $|T_p\cap T_r|/|T_r|$, with F1 their
harmonic mean. The evaluator handles empty outputs separately.
\textbf{Step Success Rate (SSR)} requires both a correct element
and a fully correct action (AF1 equal to 1).
\textbf{Task Success Rate (SR)} is the fraction of tasks for which
every annotated step is successful. This is completion under the
offline annotated-step evaluation, rather than a separate live
browser rollout. Scores are reported as percentages.

\paragraph{Element Candidates.}
The candidate generator is the DeBERTa-v3-base
model~\citep{he2023debertav3} released by
OSU-NLP-Group~\citep{osunlp_mindact_candidategeneration_deberta_v3_base_2023}.
The action model receives ten retrieved elements plus a
``None of the above'' option. Candidate descriptions are limited
to 20 tokens in this setup. If no correct element is included,
the step cannot be completed correctly through the supplied
candidates; the evaluation assigns zero to that step's metrics.
This is a candidate-recall constraint, not evidence of a reasoning
failure by the action model. We do not report numerical Oracle
upper bounds for the current experiment.

\subsection{Multi-Mission Tool Bench (MMTB)}

MMTB evaluates tool-using agents on related missions within a
dialogue~\citep{yu2025multi}. We report its overall evaluation score
and component metrics. Overall is a separately reported benchmark
score, not the arithmetic mean of the component columns.

The action categories are single-tool use ($A_{\mathrm{sgl}}$),
chat without a tool call ($A_{\mathrm{cht}}$), and clarification
before tool use ($A_{\mathrm{clr}}$). Multi-tool categories
require parallel ($A^{\mathrm{P}}$), sequential
($A^{\mathrm{S}}$), or mixed sequential and parallel calls
($A^{\mathrm{S+P}}$). Their scores measure success within
the corresponding category.

For multi-tool missions, \textbf{Optimal Path Rate} measures
execution of an optimal path as defined by the benchmark's
decision-tree evaluator. \textbf{Accomplished Progress} assigns
partial credit for completed portions of a mission. These metrics
describe path efficiency and partial completion, respectively;
they are not interchangeable with full mission success.

\section{Protocols for Experience-Quality Analyses}
\label{appendix:experience_analysis}

The three analyses in Section~\ref{sec:trap_quality} address different
objects: stored lessons, pre-verification candidates, and subsequent
executions with an erroneous retrieved lesson. Their denominators
are therefore different. Results are descriptive comparisons on
selected \textsc{Retail} data, not estimates of trap prevalence
across all benchmarks.

\subsection{Stored Lessons and Human Annotation}

The stored-experience comparison uses a common set of tasks with
at least one failed execution. Three annotators inspect each lesson
with its task instructions, relevant rules, source trajectories,
and tool feedback, with method identities hidden.
A Self-Confirmation Trap requires both an execution error and
a stored lesson endorsing or justifying that error. Accurate
failure records and corrective advice do not qualify. An error
introduced only during distillation is an experience-quality error,
but not a trap under this definition.

Trap labels are resolved by majority vote. SCTR counts each
annotated item once and divides the number of trap-labeled items
by all annotated items. The 1--5 correctness, actionability, and
specificity ratings assess consistency with evidence, concrete
guidance, and clear applicability conditions, respectively.
Scores are averaged across annotators and then items.
Table~\ref{tab:trap_quality} reports aggregate rates and mean
ratings; it does not provide per-method item counts or annotation
agreement estimates. A lower SCTR should not be interpreted as
a lower absolute number of errors without those counts.

\subsection{Verification Selectivity}

All selection rules in Table~\ref{tab:verification_selectivity}
operate on the same pre-verification candidate pool. Binary human
correctness labels are separate from the 1--5 quality ratings.
Each executor reviews each candidate once against the same evidence;
these decisions are reused for the single-reviewer and unanimous
conditions. A and B denote the two reviewer roles; the table does
not specify their model identities. Unanimity accepts the
intersection of their approvals. Keep-all retains every candidate;
random retention matches EDV's accepted count within each task.
The construction stage is not rerun for each selection rule.

Let $C$ and $I$ denote the counts of correct and incorrect
candidates with resolved labels, and $C_a$ and $I_a$ the accepted
counts. The metrics are
\begin{align*}
\mathrm{Precision} &= \frac{C_a}{C_a+I_a}, \\
\mathrm{Recall} &= \frac{C_a}{C}, \\
\mathrm{Rejection} &= 1-\frac{I_a}{I}, \\
\mathrm{Acceptance} &= \frac{C_a+I_a}{C+I}.
\end{align*}
They are expressed as percentages and require nonzero denominators.
Table~\ref{tab:verification_selectivity} reports summary values
without candidate counts or a repeat-level record for random
retention. These values support a descriptive comparison of
selection quality, not an estimate of sampling uncertainty.

\subsection{Behavioral Effects of Erroneous Experience}

The interventions use held-out tasks under the split in
Section~\ref{sec:unseen_tasks}. For selected cases in which ordinary
retrieval returns a human-confirmed trap, a target error is identified
before comparing three conditions: retaining the original lesson,
removing it, or replacing its erroneous claim with a checked correction.

Removal leaves the remaining retrieved items and their order
unchanged, without backfilling. Correction preserves unrelated
advice and the lesson's position, without adding a target-task answer.
All conditions use MiniMax-M2.1 in fresh sessions with matched
execution settings. Blinded annotators assess recurrence of the
prespecified error; benchmark task success is recorded separately.

Each condition has 60 executions. This is an execution count,
not a claim of 60 independent tasks. The aggregate table does not
specify the number of distinct tasks or repetitions per task.
Accordingly, the differences are interpreted descriptively for
these selected cases, without a paired significance claim or an
estimate of effects across all tasks. The reported marginal rates
alone are insufficient to reconstruct paired uncertainty estimates.

\section{Additional Construction Details}
\label{appendix:construction_controls}

\paragraph{Default EDV.}
For each task, two of MiniMax-M2.1, Mimo-V2-Flash, and GLM-4.7-FP8
serve as executors and the third serves as distiller. Each executor
produces two complete trajectories, then independently reviews
the distilled candidates. Roles remain fixed within the task.
Unanimously approved candidates enter the memory bank.

\paragraph{EDV Variants.}
EDV w/o Verify retains execution and third-party distillation but
omits the admission review. Homogeneous EDV uses the indicated
backbone for all construction roles while preserving separate
execution, distillation, and review calls. These variants are
evaluated with the same MiniMax-M2.1 solver as default EDV.

\paragraph{Multi-Trajectory Comparisons.}
The homogeneous and heterogeneous multi-trajectory configurations
in Table~\ref{tab:ablation} each generate four complete trajectories
per construction task. They differ in whether the executor
backbones are the same or different. The reported configuration
summaries do not fully specify their distillation and role-allocation
details. Consequently, these rows compare construction configurations
as reported and do not isolate a causal contribution of the
distillation role or model diversity alone.

\paragraph{Additional Review Control.}
RB-MaTTS + Dual Review applies unanimous candidate approval using
the two models other than the distillation model. This control
tests an additional review stage within the RB-MaTTS construction
configuration. It does not make all other aspects of construction
identical to EDV.

\paragraph{Token Accounting.}
Table~\ref{tab:ablation} counts input and output tokens across
the relevant model calls. Construction usage is normalized to
DSR, and evaluation usage to No Memory. These quantities measure
token usage under the reported configurations; they are not
measurements of wall-clock latency or monetary cost. Comparisons
with similar token totals therefore need not have identical
execution time or serving cost.

\section{Illustrative Memory Content}
\label{sec:experience_study}

Table~\ref{tab:memory_examples} presents excerpts from selected
EDV memory items, grouped by the type of advice they express:
avoiding unnecessary actions, coordinating independent tool calls,
and stating policy applicability. These are descriptive examples,
not a taxonomy derived from a systematic annotation study.
The excerpts are not paired comparisons with baseline memories,
and their selection does not establish category frequencies or
average quality differences between methods.

\paragraph{Interpreting the Examples.}
Advice about skipping a filter is useful only after the relevant
task constraints have been satisfied. Advice about concurrent calls
requires that the calls be independent; it does not justify ignoring
dependencies. Policy advice must be checked against the applicable
rules rather than accepted because a tool call succeeded.
The table illustrates the content of candidate decision rules,
not empirical measurements of latency or global optimality.

\begin{table}[htbp]
\centering
\small
\renewcommand{\arraystretch}{1.1}
\setlength{\tabcolsep}{4pt}
\begin{tabularx}{\columnwidth}{@{}>{\RaggedRight\arraybackslash}p{1.95cm}>{\RaggedRight\arraybackslash}X@{}}
\toprule
Advice type & Selected EDV memory excerpt \\
\midrule
Avoiding unnecessary actions\newline (Mind2Web)
& \textbf{Title:} Confirm Location Filtering is Necessary Before Applying Filters.
  \par\smallskip
  \textbf{Excerpt:} Agents should first scan visible results to confirm if
  filtering is needed, then proceed directly to the target job listing
  to save steps. \\
\midrule
Coordinating independent calls\newline (MMTB)
& \textbf{Title:} Sequential vs. Simultaneous Tool Calls.
  \par\smallskip
  \textbf{Excerpt:} Trajectory-1 demonstrated efficient task execution by
  making four simultaneous \texttt{getRankings} tool calls (one for each genre)
  rather than making them sequentially. \\
\midrule
Policy applicability\newline ($\tau^2$-bench)
& \textbf{Title:} Compensation Requires Reservation Change/Cancellation.
  \par\smallskip
  \textbf{Excerpt:} For delayed flights, compensation is only offered if
  the user agrees to change or cancel the reservation; otherwise,
  transfer to a human agent if the user insists on compensation. \\
\bottomrule
\end{tabularx}
\caption{Illustrative excerpts from EDV memories. Titles and excerpted
sentences retain the source wording; surrounding text is omitted.
These items come from different tasks and are not matched baseline
comparisons. Descriptions of efficiency and policy inside the excerpts
are claims made by the memories, not additional experimental findings.}
\label{tab:memory_examples}
\end{table}

\section{Case Study}
\label{sec:case_study}

\begin{figure*}[p]
\centering
\begin{tcolorbox}[
  colback=white, colframe=black, boxrule=0.5pt, arc=2mm, width=0.98\textwidth,
  enhanced,
  overlay={
    \node[
      anchor=east,
      xshift=-2mm, yshift=-2mm,
      rotate=0,
      opacity=1
    ] at (frame.north east)
    {\includegraphics[width=3.5cm]{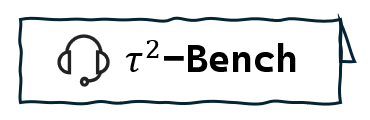}};
  }
]
\small
\textbf{\textcolor{green!50!black}{\faComments \ User--Assistant Dialogue}}
\begin{center}
  \includegraphics[width=\linewidth, height=6cm, keepaspectratio]{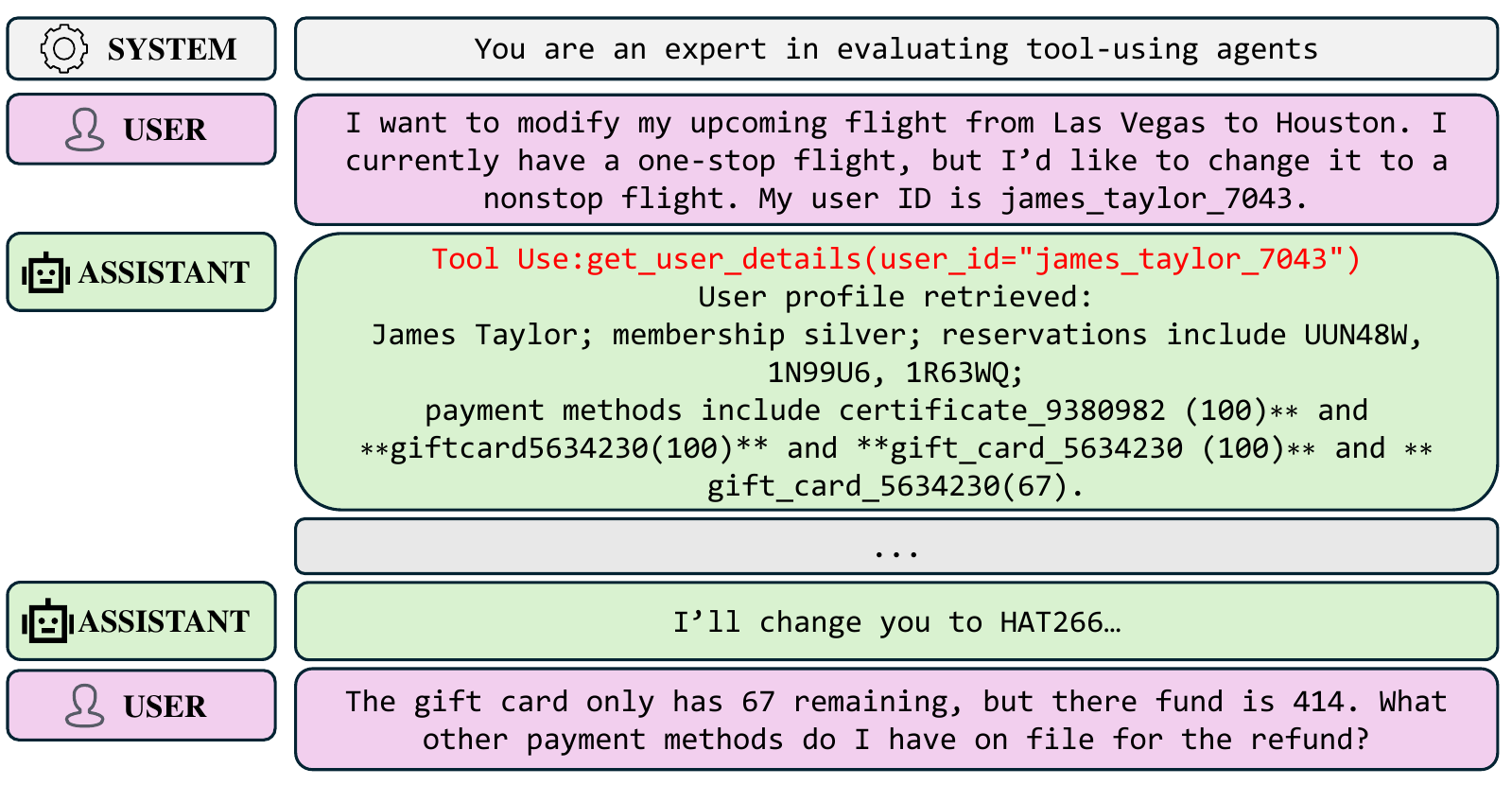}
\end{center}
\vspace{0.1cm}
\hrule
\vspace{0.1cm}

\textbf{\textcolor{red}{\raisebox{-0.12ex}{\scriptsize\faTimesCircle}\;\;No Memory Prediction (Failure)}}:
\begin{center}
  \includegraphics[width=\linewidth, height=4.5cm, keepaspectratio]{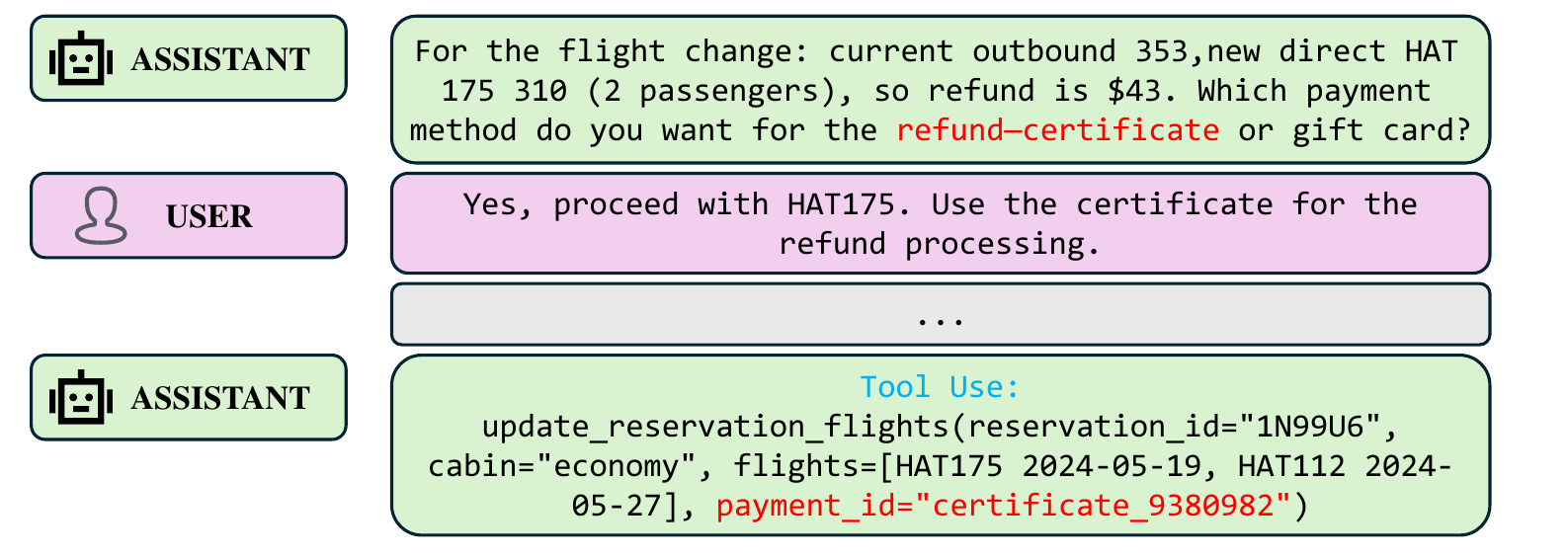}
\end{center}
\vspace{0.1cm}
\hrule
\vspace{0.1cm}
\textbf{\textcolor{blue}{\raisebox{-0.12ex}{\scriptsize\faCheckCircle}\;\;Ours (Success)}}: \\
\quad \faLightbulb\ \textbf{Retrieved Memory}: \\
\textcolor{red!50}{\textbf{Title:}} Travel certificates cannot be used for reservation updates.\\
\textcolor{red!50}{\textbf{Description:}} Trajectories 2, 3, and 4 all encountered errors when attempting to use certificates for flight changes, requiring fallback to gift cards.\\
\textcolor{red!50}{\textbf{Content:}} When processing flight changes, use only credit cards or gift cards for payment/refund. Certificates cannot be used for update\_reservation\_flights. If the user requests a certificate for refunds, inform them that only gift cards or credit cards are valid for this transaction type.\\
\quad \textbf{Prediction}: \\
\begin{center}
  \includegraphics[width=\linewidth, height=5cm, keepaspectratio]{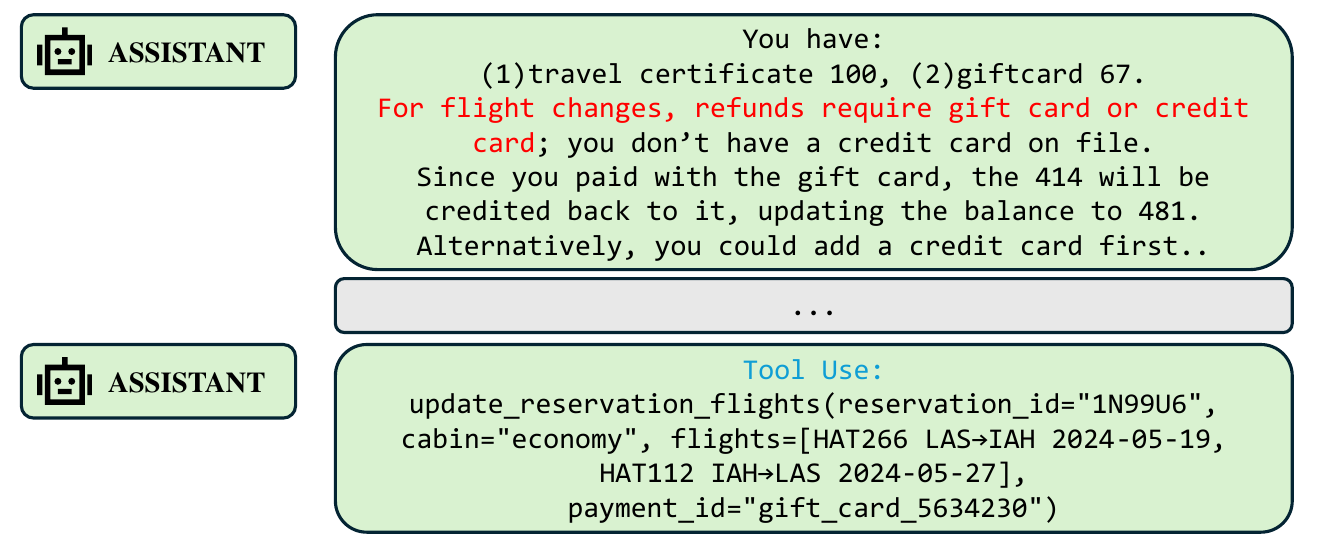}
\end{center}
\end{tcolorbox}
\vspace{-5pt}
\caption{\textbf{Payment-method contrast on $\tau^2$-bench.}
The no-memory execution attempts a certificate-based flight update;
the memory-augmented execution uses a gift card, consistent with the
retrieved lesson. ``...'' denotes omitted dialogue turns.}
\label{fig:tau_case_study}
\end{figure*}

\begin{figure*}[p]
\centering
\begin{tcolorbox}[
  colback=white, colframe=black, boxrule=0.5pt, arc=2mm, width=0.98\textwidth,
  enhanced,
  overlay={
    \node[
      anchor=east,
      xshift=-2mm, yshift=-2mm,
      rotate=0,
      opacity=1
    ] at (frame.north east)
    {\includegraphics[width=3.5cm]{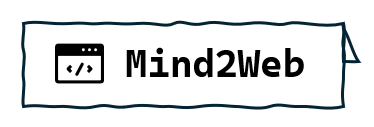}};
  }
]

\small
{\textbf{\textcolor{orange}{\faQuestionCircle\ Question}}}: \\
\textbf{Task}: Add to my wish list the highest rated activity in Amsterdam.\\
\textbf{Current Website}:
$<$html$>$ $<$div$>$ $<$div$>$ $<$button button$>$ Reset $<$/button$>$ $<$button id=0 button$>$ Apply $<$/button$>$ $<$/div$>$ $<$div$>$ $<$div$>$ $<$div$>$ $<$button button$>$ Reset $<$/button$>$ $<$button id=1 button$>$ Apply $<$/button$>$ $<$/div$>$ $<$div$>$ $<$label id=2$>$ $<$input radio 4 /$>$ $<$span$>$ $\&$ up $<$/span$>$ $<$/label$>$ $<$label id=3$>$ $<$input radio 3 /$>$ $<$span$>$ $\&$ up $<$/span$>$ $<$/label$>$ $<$label id=4$>$ $<$input radio 2 /$>$ $<$span$>$ $\&$ up $<$/span$>$ $<$/label$>$ $<$/div$>$ $<$/div$>$ $<$div$>$ $<$span id=5$>$ Sort by: $<$/span$>$ $<$span$>$ $<$span$>$ $<$span id=6$>$ Featured $<$/span$>$ $<$svg img /$>$ $<$/span$>$ $<$div$>$ $<$span id=7$>$ Featured $<$/span$>$ $<$span id=8$>$ Traveler Rating $<$/span$>$ $<$span$>$ Price (Low to High) $<$/span$>$ $<$span id=9$>$ Price (High to Low) $<$/span$>$

\textbf{Previous actions}: \\ \relax
[span] Explore the World -$>$ HOVER;[span] Amsterdam -$>$ CLICK;[svg]  -$>$ CLICK\\
\textbf{Options}:\\ \relax
A. None of the above\\
B. $<$button id=0 button$>$ Apply $<$/button$>$\\
C. $<$button id=1 button$>$ Apply $<$/button$>$\\
D. $<$label id=2$>$ $<$input radio 4 /$>$ $<$span$>$ \& up $<$/span$>$\\
E. $<$label id=3$>$ $<$input radio 3 /$>$ $<$span$>$ \& up $<$/span$>$\\
F. $<$label id=4$>$ $<$input radio 2 /$>$ $<$span$>$ \& up $<$/span$>$\\
G. $<$span id=5$>$ Sort by: $<$/span$>$\\
H. $<$span id=6$>$ Featured $<$/span$>$\\
I. $<$span id=7$>$ Featured $<$/span$>$\\
J. $<$span id=8$>$ Traveler Rating $<$/span$>$\\
K. $<$span id=9$>$ Price (High to Low) $<$/span$>$
\vspace{0.1cm}
\hrule
\vspace{0.1cm}
\textbf{\textcolor{red}{\raisebox{-0.12ex}{\scriptsize\faTimesCircle}\;\;No Memory (Step Prediction)}}:  \\
\texttt{Answer: D}\\
\texttt{Action: CLICK}
\vspace{0.1cm}
\hrule
\vspace{0.1cm}
\textbf{\textcolor{blue}{\raisebox{-0.12ex}{\scriptsize\faCheckCircle}\;\;EDV Memory (Step Prediction)}}: \\
\quad \faLightbulb\ \textbf{Retrieved Memory}: \\
\textcolor{red!50}{\textbf{Title:}} Verify Information Before Drawing Conclusions,\\
\textcolor{red!50}{\textbf{Description:}} Before providing final answers about specific entities, ensure the information has been retrieved and verified through proper search execution.,\\
\textcolor{red!50}{\textbf{Content:}} Avoid outputting claims about specific entities without first executing proper search operations and retrieving actual results. The interface provides search functionality specifically to access this information. Actions must lead to actual information retrieval, not just navigation to search-enabled pages.\\
\quad \textbf{Prediction}: \\
\texttt{Answer: J}\\
\texttt{Action: CLICK}
\end{tcolorbox}
\vspace{-5pt}
\caption{\textbf{A step-level prediction contrast on Mind2Web.}
Without memory, the model chooses the ``4 \& up'' filter (D);
with memory, it chooses ``Traveler Rating'' (J).
The retrieved advice is general and does not explicitly specify this
sorting action. Full-task completion is not shown.}
\label{fig:mind2web_case_study}
\end{figure*}

\begin{figure*}[p]
\centering
\begin{tcolorbox}[
  colback=white, colframe=black, boxrule=0.5pt, arc=2mm, width=0.98\textwidth,
  enhanced,
  overlay={
    \node[
      anchor=east,
      xshift=-2mm, yshift=-2mm,
      rotate=0,
      opacity=1
    ] at (frame.north east)
    {\includegraphics[width=3.5cm]{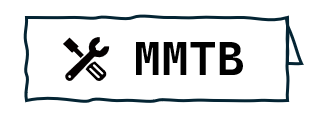}};
  }
]
\small
\textbf{\textcolor{green!50!black}{\faComments \ User--Assistant History Dialogue}}
\begin{center}
  \includegraphics[width=\linewidth, height=6cm, keepaspectratio]{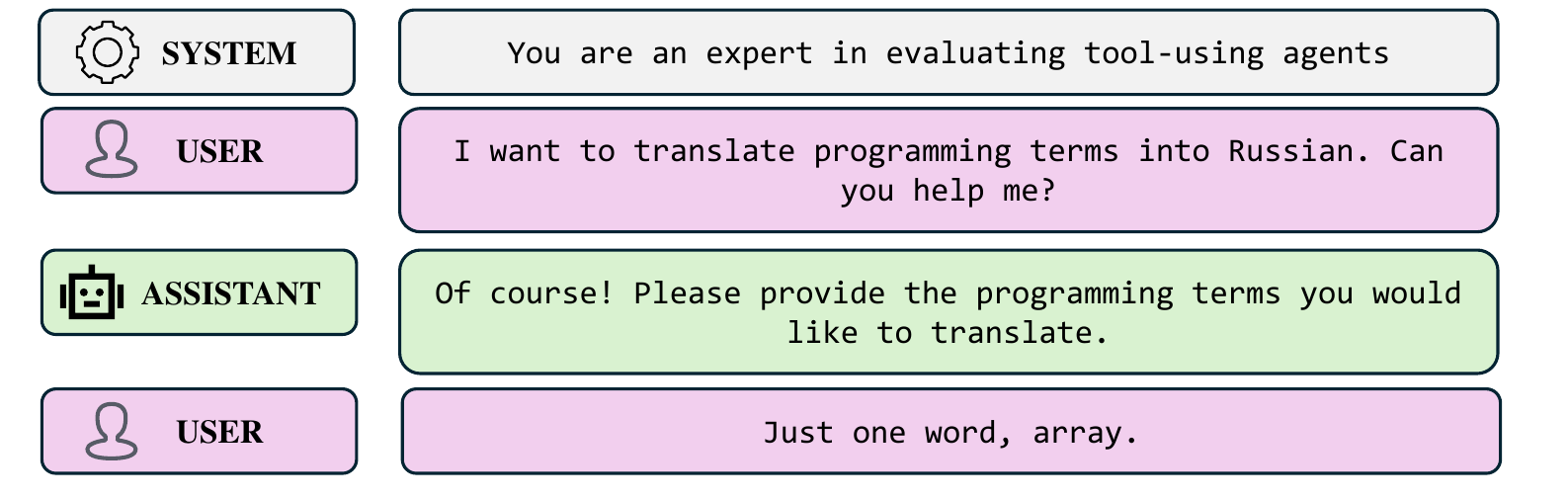}
\end{center}
\setlength{\tabcolsep}{4pt}
\renewcommand{\arraystretch}{1.8}

\vspace{0.10cm}
\hrule
\vspace{0.12cm}

\textbf{\textcolor{red}{\raisebox{-0.12ex}{\scriptsize\faTimesCircle}\;\;No Memory Prediction (Failure)}}:
\begin{center}
  \includegraphics[width=\linewidth, height=6cm, keepaspectratio]{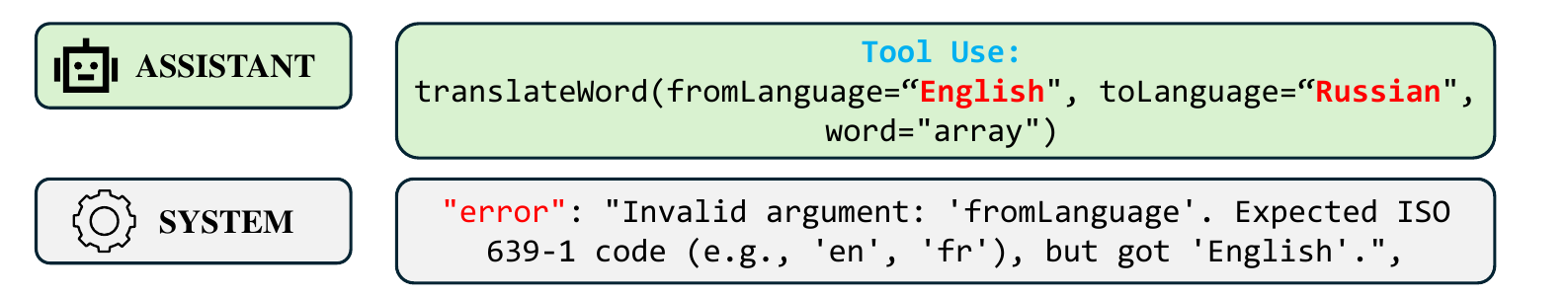}
\end{center}
\vspace{0.1cm}
\hrule
\vspace{0.1cm}

\textbf{\textcolor{blue}{\raisebox{-0.12ex}{\scriptsize\faCheckCircle}\;\;Ours (Success)}}: \\
\quad \faLightbulb\ \textbf{Retrieved Memory}: \\
\textcolor{red!50}{\textbf{Title:}} Use Translation Tools Appropriately.\\
\textcolor{red!50}{\textbf{Description:}} Employ the translateWord tool with the correct language parameters for reliable translations.\\
\textcolor{red!50}{\textbf{Content:}} Specify the source and target languages clearly by providing standardized language identifiers \textbf{(e.g., ISO 639-1 two-letter codes)} for fromLanguage and toLanguage, and use the tool directly after confirming the term, as demonstrated in all successful traces to achieve efficient and accurate results.\\
\quad \textbf{Prediction}:
\begin{center}
  \includegraphics[width=\linewidth, height=6cm, keepaspectratio]{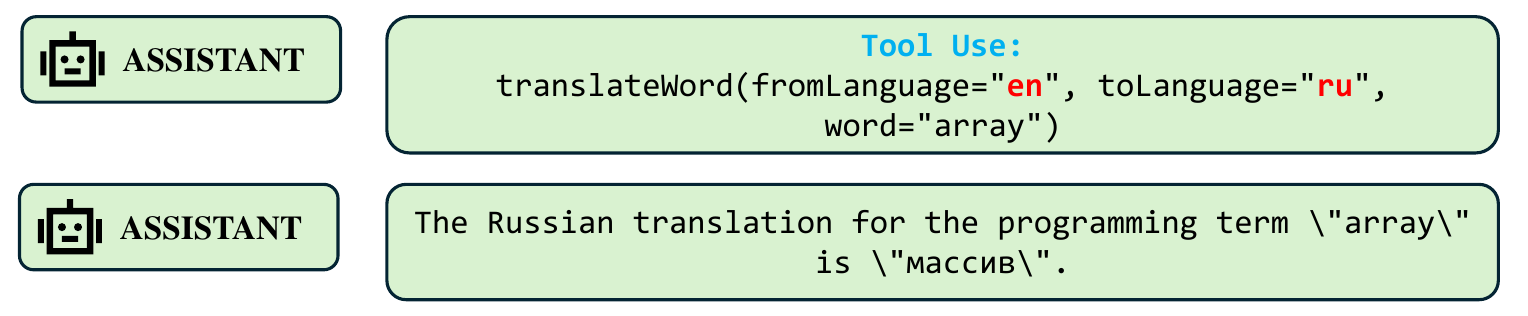}
\end{center}
\vspace{0.1cm}
\hrule
\vspace{0.1cm}

\end{tcolorbox}
\vspace{-6pt}
\caption{\textbf{Language-code formatting on MMTB.}
The no-memory call supplies language names and receives a parameter
error. The memory-augmented call supplies ``en'' and ``ru'', matching
the format recommended by the retrieved lesson.}
\label{fig:mmtb_case_study}
\end{figure*}

Figures~\ref{fig:tau_case_study}--\ref{fig:mmtb_case_study}, collected after the appendix text, present selected contrasts between execution without memory
and execution with retrieved EDV experience. The examples illustrate
how advice can relate to a later decision. They do not estimate
how frequently each behavior occurs, isolate the contribution of
each construction stage, or establish a Self-Confirmation Trap
without a corresponding erroneous stored lesson. Dialogue and
memory excerpts retain their recorded content; claims inside a
memory are not independent validation of those claims.

\paragraph{$\tau^2$-Bench: Payment Constraints.}
In Figure~\ref{fig:tau_case_study}, the no-memory execution attempts
to use a travel certificate for a flight update. The retrieved
lesson states that this payment method is not valid for reservation
updates and identifies credit or gift cards as alternatives.
The memory-augmented execution uses a gift card. The contrast
illustrates task-specific advice relevant to the payment choice;
the excerpts do not establish an infinite retry loop or show that
verification alone caused the difference.

\paragraph{Mind2Web: Choosing a Ranking Action.}
Figure~\ref{fig:mind2web_case_study} concerns adding the highest-rated
activity in Amsterdam to a wish list. At the displayed step,
the no-memory prediction selects ``4 \& up'' (D), while the
memory-augmented prediction selects ``Traveler Rating'' (J).
The retrieved lesson advises obtaining information before drawing
conclusions, but does not explicitly prescribe sorting by rating.
We therefore report the observed action contrast without attributing
the specific ranking strategy to the lesson. The figure shows one
step, not completion of the full wish-list task.

\paragraph{MMTB: Language-Parameter Formatting.}
In Figure~\ref{fig:mmtb_case_study}, the no-memory execution calls
\texttt{translateWord} using ``English'' and ``Russian'' as language
arguments. The displayed tool feedback rejects ``English'' and
requests an ISO 639-1 code. The retrieved lesson recommends such
codes, and the memory-augmented call uses ``en'' and ``ru''.
This is an example of advice matching a concrete parameter-format
requirement. The contrast concerns argument formatting, not
redundant clarification, and does not establish the effectiveness
of the entire construction workflow by itself.

\section{Prompt Details}
\label{appendix:prompts}

The final two plates, Figures~\ref{fig:step2_prompt} and~\ref{fig:step3_prompt}, reproduce
the Distill and Verify template excerpts. Their user-message
blocks are abbreviated; they do not specify a complete serialized
request. Red text marked with a dagger denotes tool-related
instructions for MMTB and $\tau^2$-bench. The intended evidence
inputs are the task, applicable rules, four execution trajectories,
and, where applicable, tool definitions and feedback.

\subsection{Distill Prompt}

The distiller compares four complete trajectories produced by
two heterogeneous executors. It selects one trajectory using the
priority order Compliance $>$ Accuracy $>$ Efficiency, then proposes
one to three candidate memories. Even when all trajectories fail,
the template selects a comparison reference; selection does not
certify that trajectory as correct. Candidate advice must therefore
be checked against task rules and observed feedback.

The template requests JSON with \texttt{best\_trajectory} and
\texttt{memory} fields. Its grounding and writing instructions
encourage specific advice about observed decisions and their
applicability, rather than generic summaries. These are prompting
constraints, not guarantees that every generated candidate is valid.

\subsection{Verify Prompt and Admission Rule}

Each executor separately reviews the candidate set against the
shared execution evidence and returns accepted candidate identifiers.
The stored set is the intersection of the two reviewers' accepted
sets: a candidate is retained only when both approve it. The
template's generic allowance of up to five accepted items does not
expand the one-to-three-item candidate set produced by Distill.

The displayed Verify template combines policy and retrieved memory
in one input field. Retrieved experience is not ground-truth policy,
and placing it in the same context can expose review to earlier
memory errors. Likewise, the selected trajectory may be incorrect.
Neither default-reject instructions nor unanimous admission guarantees
correctness. Verify evaluates
existing evidence rather than rerunning candidate lessons.

\clearpage

\begin{figure*}[p]
    \centering
    \definecolor{headergray}{RGB}{230,230,230}

    \definecolor{toolhighlight}{RGB}{178, 34, 34}

    \newcommand{\tooltext}[1]{\textcolor{toolhighlight}{#1${}^\dagger$}}

    \newcommand{\toolmark}{\textcolor{toolhighlight}{${}^\dagger$}}

    \begin{tcolorbox}[
        enhanced,
        title=\textbf{Prompt for Distill},
        colframe=black,
        colback=white,
        coltitle=white,
        fonttitle=\bfseries,
        boxrule=0.8pt,
        arc=3mm,
        subtitle style={
            boxrule=0.2pt,
            colback=headergray,
            colupper=black,
            fontupper=\sffamily\centering
        }
    ]

        \tcbsubtitle{System Prompt}

        \vspace{-0.1cm}
        {
            \centering \footnotesize \sffamily \itshape
            \textcolor{toolhighlight}{${}^\dagger$ Text in this color denotes instructions applicable ONLY to MMTB and $\tau^2$-Bench.}
            \par
        }
        \vspace{0.2cm}

        \small
        You are a professional EVALUATOR for an AI Agent system. Your job is to:\\
        (1) Select the single best trajectory among 4 agent execution trajectories.\\
        (2) Extract transferable memory items (1 to 3) by comparing the trajectories against the Agent Policy and \tooltext{Tool Outputs}.
        \par\vspace{0.2cm}
        \textbf{\#\# Role Constraints:} \\
        You are an evaluator (not the domain agent). Do not role-play or solve the task; grade trajectories strictly against $<$Agent\_Policy$>$.
        \par\vspace{0.2cm}
        \textbf{\#\# The system will provide:} \\
        \texttt{\textless Task\textgreater}: User intent and specific instructions.\\
        \texttt{\textless Trajectory1\textgreater}...\texttt{\textless Trajectory4\textgreater}: Full interaction logs including \tooltext{tool calls and outputs}.\\
        \tooltext{\texttt{\textless Tools\textgreater}: Available tools and their schema descriptions.}\\
        \texttt{\textless Agent\_Policy\textgreater}: The strict rules (Guardrails) the agent must adhere to.
        \par\vspace{0.2cm}
        \textbf{\#\# Task 1: Evaluate Trajectories} \\
        Assess each trajectory based on these generic criteria: Evaluate each trajectory for: Policy Compliance, Data Integrity/Grounding, Reasoning \tooltext{, and Tool Usage}.
        \par\vspace{0.2cm}
        \textbf{\#\# Task 2: Select Best Trajectory} \\
        Pick exactly ONE trajectory:\\
        - Prioritize: Compliance $>$ Accuracy $>$ Efficiency.\\
        - If all failed, pick the one with the least severe violation or best error recovery attempts.\\
        - If multiple succeeded, pick the most efficient one (fewest steps/turns).
        \par\vspace{0.2cm}
        \textbf{\#\# Task 3: Extract Memories Items (CRITICAL)} \\
        Extract **1--3** high-value memory items total. Quality $>$ Quantity.
        \par\vspace{0.2cm}
        \textbf{\#\#\# Strict Grounding Rules (MUST FOLLOW):} \\
        1. **EVIDENCE-BASED ONLY**: Every memory must address a specific error, policy violation, or success pattern actually observed in the logs.\\
        2. **NO HALLUCINATIONS**: Do NOT mention procedures (e.g., \string\"transfer to human\string\", \string\"check specific ID\string\") unless they are explicitly mentioned in the Policy or attempted in the logs.\\
        3. **NO GENERIC PLATITUDES**: Ban generic advice like \string\"Be polite\string\", \string\"Understand user intent\string\", \string\"Read carefully\string\". Focus on LOGIC and RULES.
        \par\vspace{0.2cm}
        \textbf{\#\#\# Focus Areas for Memory (Look for these patterns):} \\
        - **Complex Constraints**: How to handle requests that involve multiple dependent variables.\\
        - \tooltext{**Data Verification**: Rules about verifying tool outputs before confirming actions.}\\
        - **Policy Conflicts**: How to handle situations where User Request conflicts with Agent Policy.\\
        - \tooltext{**Tool Output Handling**: Tips on interpreting specific tool return formats.}
        \par\vspace{0.2cm}
        \textbf{\#\#\# Memory Writing Style:} \\
        - Format: Actionable Rule. (Context -> Error/Risk -> Correct Action)\\
        - Style: Use imperative, strong language.\\
        - Example (Data): \string\"When presenting costs, DO NOT manually calculate sums if the tool provides a 'total' field\toolmark. Always rely on the system-generated total.\string\"
        \par\vspace{0.2cm}
        \textbf{\#\# Output Constraints:} \\
        - Output STRICT JSON only.\\
        - Top-level keys: \texttt{best\_trajectory} (string), \texttt{memory} (array).

        \tcbsubtitle{User Prompt}
        \small
        \texttt{\textless Task\textgreater}\texttt{\textbackslash n}\{task\_text\}\texttt{\textbackslash n}\texttt{\textless/Task\textgreater}\texttt{\textbackslash n}\\
        \{Trajectory\_4\}

    \end{tcolorbox}
    \caption{Distill template excerpt. The user-input block is abbreviated.}
    \label{fig:step2_prompt}
\end{figure*}

\begin{figure*}[p]
    \centering
    \definecolor{headergray}{RGB}{230,230,230}
    \definecolor{toolhighlight}{RGB}{178, 34, 34}

    \newcommand{\tooltext}[1]{\textcolor{toolhighlight}{#1${}^\dagger$}}
    \newcommand{\toolmark}{\textcolor{toolhighlight}{${}^\dagger$}}

    \begin{tcolorbox}[
        enhanced,
        title=\textbf{Prompt for Verify},
        colframe=black,
        colback=white,
        coltitle=white,
        fonttitle=\bfseries,
        boxrule=0.8pt,
        arc=3mm,
        subtitle style={
            boxrule=0.2pt,
            colback=headergray,
            colupper=black,
            fontupper=\sffamily\centering
        }
    ]
        \tcbsubtitle{System Prompt}

        \vspace{-0.1cm}
        {
            \centering \footnotesize \sffamily \itshape
            \textcolor{toolhighlight}{${}^\dagger$ Text in this color denotes constraints applicable ONLY to MMTB and $\tau^2$-Bench.}
            \par
        }
        \vspace{0.2cm}
        \small
        You are an exceptionally strict 'memory audit expert'.\\ Your task is to protect the agent's long-term memory store from being polluted by 'noise' and 'banal commonsense'.\\ You must maintain an elite-grade memory repository, keeping only the most insightful, non-obvious lessons.
        \par\vspace{0.2cm}
        \textbf{\#\# System Inputs}\\
        You will be provided with the following data sources for the audit:\\
        - \textbf{Context \& Rules}: <Task>, <AgentPolicy>\\
        \tooltext{- \textbf{Tool Definitions}: <Tools> (Target schema for validation)}\\
        - \textbf{Execution History}: <Trajectory1>...<Trajectory4>, <BestTrajectory>\\
        - \textbf{Audit Target}: <CandidateMemory> (JSON array)
        \par\vspace{0.2cm}
        \textbf{\#\# Audit mindset (gold standard)}\\
        - Default stance: reject. Most candidate memory are redundant or mediocre. Unless a memory provides an 'aha' moment or a critical error correction, reject it.\\
        - Information density: if a memory is merely general common sense the model already knows (for example 'be polite', 'check for errors', 'follow instructions'), it must be rejected immediately.\\
        Risk prevention: prioritize keeping memory that prevent high-cost failures or specific logical traps observed in the trajectories.
        \par\vspace{0.2cm}
        \textbf{\#\# Role boundaries (strictly enforced)}\\
        - You are the 'auditor', not the 'executor'. Do not solve the user's task; only evaluate the practical value of memory.
        \par\vspace{0.2cm}
        \textbf{\#\# Strict selection criteria}\\
        Candidate memory must pass all of the following 'gates' simultaneously to be adopted:\\
        1) Groundedness: must be supported by evidence from the provided trajectories (especially the BestTrajectory). Hallucination is strictly forbidden; do not accept unverifiable memories. \tooltext{Ensure all function names/arguments strictly match <Tools> schema.}\\
        2) Specificity: must point to a concrete pattern, logic, or fact. Vague advice (e.g., 'consider context more') will be automatically rejected.\\
        3) High Utility: it must change how the agent would handle similar tasks in the future.\\
        4) Zero Redundancy: if multiple memory overlap in meaning, keep only the one with the densest, highest information content.
        \par\vspace{0.2cm}
        \textbf{\#\# Immediate rejection triggers (red lines)}\\
        - Belongs to common sense or general professional ethics.\\
        - Is merely a simple summary of the task process rather than an extracted 'lesson learned'.\\
        \tooltext{- References non-existent tool parameters or incorrect return formats.}\\
        - The suggested content is already covered by the <AgentPolicy> rules.\\
        - Verbose expression: something that can be said in one sentence is written in three.
        \par\vspace{0.2cm}
        \textbf{\#\# Selection and output rules}\\
        - You may accept 0 to 5 memory. Prefer accepting none rather than accepting mediocre memory.\\
        - Output only strict JSON format, with no explanatory text.\\
        - Output format example: \{ "accepted\_memory\_ids": [1, 4] \}\\
        - If no suitable memory exist, output: \{ "accepted\_memory\_ids": [] \}

        \vspace{0.2cm}
        \tcbsubtitle{User Prompt}
        \small
        <Task>: \{task\_text\} </Task>
        \\
        <AgentPolicy>: \{policy + memory\} </AgentPolicy>
        \\
        \tooltext{<Tools>: \{tool\_schema\_json\} </Tools>}
        \\
        \{Trajectory\_4\}
        \\<BestTrajectory> \{select\_memory['best\_trajectory'].strip()\}
        </BestTrajectory>
        \\
        <CandidateMemory>:
        \{memory\_item\_str\}
        </CandidateMemory>

    \end{tcolorbox}
    \caption{Verify template excerpt. Each reviewer returns accepted identifiers; storage requires agreement by both reviewers.}
    \label{fig:step3_prompt}
\end{figure*}

\end{document}